\documentclass[sigconf]{acmart}

\usepackage{booktabs} 
\usepackage{tablefootnote}
\usepackage[font=small]{caption}

\newcommand{\eg}{\emph{e.g.}}

\newcommand{\etal}{\emph{et~al.}}

\setcopyright{none}

\acmYear{2017} 
\acmConference{Accepted to CIKM'17 }{November 6--10, 2017}{Singapore, Singapore}
\acmPrice{}
\acmDOI{10.1145/3132847.3133104}
\acmISBN{978-1-4503-4918-5/17/11}

\begin{document}
\title{Language Modeling by Clustering with Word Embeddings\\for Text Readability Assessment}

\author{Miriam Cha}
\affiliation{Harvard University}
\email{miriamcha@fas.harvard.edu}

\author{Youngjune Gwon}
\affiliation{Harvard University}
\email{gyj@eecs.harvard.edu}

\author{H.~T.~Kung}
\affiliation{Harvard University}
\email{kung@harvard.edu}

\begin{abstract}
We present a clustering-based language model using word embeddings for text readability prediction. Presumably, an Euclidean semantic space hypothesis holds true for word embeddings whose training is done by observing word co-occurrences. We argue that clustering with word embeddings in the metric space should yield feature representations in a higher semantic space appropriate for text regression. Also, by representing features in terms of histograms, our approach can naturally address documents of varying lengths. An empirical evaluation using the Common Core Standards corpus reveals that the features formed on our clustering-based language model significantly improve the previously known results for the same corpus in readability prediction. We also evaluate the task of sentence matching based on semantic relatedness using the Wiki-SimpleWiki corpus and find that our features lead to superior matching performance. 
\end{abstract}

%
%

\ccsdesc[500]{Information systems~Clustering}

\keywords{Readability assessment, clustering-based language model}

\maketitle

\section{Introduction}
Predicting reading difficulty of a document is an enduring problem in natural language processing (NLP). Approaches based on shallow-length features of text date back to 1940s \cite{flesch48}. Remarkably, they are still being used and extended with more sophisticated techniques today. In this paper, we use word embeddings to compose semantic features that are presumably beneficial for assessing text readability. Encouraged by the recent literature in applying language models for better prediction, we aim to build a clustering-based language model using word vectors learned from corpora. The resulting model is expected to reveal semantics at a higher level than word embeddings and provide discriminative features for text regression. 

As pioneering work in text difficulty prediction, Flesch \cite{flesch48} explored on shallow-length features computed by averaging the number of words per sentence and the number of syllables per word. The intent was to capture sentence complexity with the number of words, and word complexity with the number of syllables. Chall \cite{chall} claimed the reading difficulty as a linear function of shallow-length features. Kincaid \cite{kincaid} introduced a linear weighting scheme that became the most common measure of reading difficulty based on shallow-length features. More sophisticated algorithms that measure semantics by word frequency counts and syntax from sentence length \cite{stenner} and language modeling \cite{collins} have shown significant performance gains over classical methods. 

Modern approaches treat text difficulty prediction as a discriminative task. Schwarm~\etal~\cite{schwarm} presented text regression based on support vector machine (SVM). Peterson~\etal~\cite{petersen} used both SVM classification and regression for improvement. NLP researchers went beyond the shallow features and looked into learning complex lexical and grammatical features. Flor~\etal~\cite{flor2013} proposed an algorithm that measures lexical complexity from word usage. Vajjala~\etal~\cite{vajjala2014} formulated semantic and syntactic complexity features from language modeling, which resulted some improvement. Class-based language models, trained on the conditional probability of a word given the classes of its previous words, were commonly used in the literature \cite{turian2010,botha2014}. Brown clustering \cite{brown}, a popular class-based language model, can learn hierarchical clusters of words by maximizing the mutual information of word bigrams. 

Our text learning is founded on word embeddings. Bengio~\etal~\cite{bengio} proposed an early neural embedding framework. Mikolov~\etal~\cite{mikolov} introduced the Skip-gram model for efficient training with large unstructured text, and Paragraph Vector \cite{doc2vec} and character $n$-grams \cite{fasttext}, all of which we use for our implementation in this paper, followed on. Most word embedding algorithms build on the distributional hypothesis \cite{harris1954} that word co-occurrences imply similar meaning and context. Word embeddings span a high-dimensional semantic space where the Euclidean distance between word vectors measures their semantic dissimilarity \cite{hashimoto2016}. 

Under the Euclidean semantic space hypothesis, we argue that clustering of word vectors should unveil a clustering-based language model. In particular, we propose two clustering methods to construct language models using Brown clustering and $K$-means with word vectors. Our methods are language-independent and data-driven, and we have empirically validated their superior performance in text readability assessment. Specifically, our experiment on the Common Core Standards corpus reveal that the language model learned by $K$-means significantly improves readability prediction against contemporary approaches using the lexical and syntactic features. In another experiment with the Wiki-SimpleWiki corpus, we show that our features can correctly identify sentence pairs of the similar meaning but written in different vocabulary and grammatical structure.

For text with easy readability, difference in reading difficulty is resulted from different document length, sentence structure, and word usage. For documents at higher reading levels, however, features with richer linguistic context about domain, grammar, and style are known to be more relevant. For example, based on shallow features, ``To be or not to be, that is the question'' would likely be considered easier than ``I went to the store and bought bread, eggs, and bacon, brought them home, and made a sandwich.'' Therefore, we need to capture all semantic, lexical, and grammatical features for distinguishing documents at all levels. 

We organize the rest of this paper as follows. In Section 2, we describe our approach centered around neural word embedding and probabilistic language modeling. We will explain each component of our approach in detail. Section 3 presents our experimental methodology for evaluation. We will also discuss the empirical results. Section 4 concludes the paper.  

\begin{figure}[t]
\centering
\includegraphics[width=.3\textwidth]{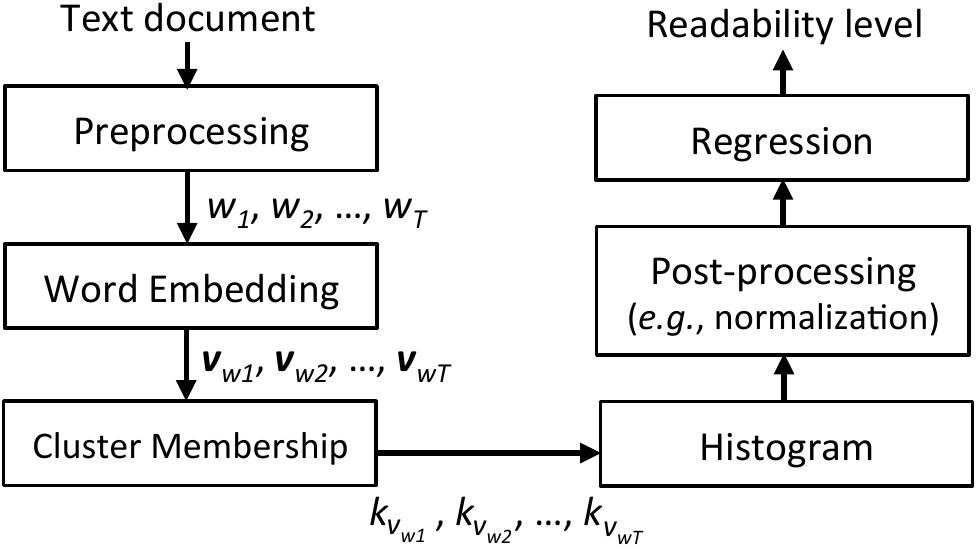}
\caption{Our system pipeline}
\label{fig:ours}
\end{figure} 

\section{Approach}
We review embedding schemes, clustering algorithms, and regression method used in the paper, and describe our overall  pipeline. 

\subsection{Word embeddings}
\textbf{Skip-gram.} Mikolov~\etal~\cite{mikolov} proposed the Skip-gram method based on a neural network that maximizes 
\begin{equation}
\label{eq:skipgram}
\small
\frac{1}{T} \sum_{t=1}^{T} \sum_{-c \leq j \leq c, j\neq0} \log\,p(w_{t+j}|w_t) 
\end{equation} 
where the training word sequence $w_1, w_2,\dots, w_T$ has a length $T$. With $w_t$ as the center word, $c$ is the training context window. The conditional probability can be computed with the softmax function 
\begin{equation}
\label{eq:sm}
p(w_{t+j}|w_t) = \frac{e^{s(w_{t},w_{t+j})}}{\sum_{w'}e^{s(w_{t},w')}} 
\end{equation} 
with the scoring function $s(w_{t},w_{t+j}) = \mathbf{v}_{w_t}^\top \cdot \mathbf{v}_{w_{t+j}}$. The embedding $\mathbf{v}_{w_t}$ is a vector representation of the word $w_t$. 

\noindent\textbf{Bag of character $n$-grams.} Bojanowski~\etal~\cite{fasttext} proposed an embedding method by representing each word as the sum of the vector representations of its \emph{character} $n$-grams. To capture the internal structure of words, a different scoring function is introduced
\begin{equation}
\label{eq:ft}
\small
s(w_t,w_{t+j}) = \sum_{g \in G_{w_t}} \mathbf{z}_g^\top \cdot \mathbf{v}_{w_{t+j}}. 
\end{equation}
Here, $G_{w_t}$ is the set of $n$-grams in $w_t$. A vector representation $\mathbf{z}_g$ is associated to each $n$-gram $g$. This approach has an advantage in representing unseen or rare words in corpus. If the training corpus is small, character $n$-grams can outperform the Skip-gram (of words) approach.

\subsection{Paragraph embedding}
\textbf{Distributed bag-of-words.} While Skip-gram and character $n$-grams can embed a word into a high-dimensional vector space, we eventually need to compute a feature vector for the whole document. Le~\etal~\cite{le} introduced Paragraph Vector that learns a fixed-length vector representation for variable-length text such as sentences and paragraphs. The distributed bag-of-words version of Paragraph Vector has the same architecture as the Skip-gram model except that the input word vector is replaced by a paragraph token. 

\subsection{Clustering}
\textbf{Brown clustering.} Brown~\etal~\cite{brown} introduced a hierarchical clustering algorithm that maximizes the mutual information of word bigrams. The probability for a set of words $w_1, w_2,\dots, w_T$ can be written as 
\begin{equation}
\label{eq:brown}
\small
\prod^{T}_{t=1}p(w_{t}|C(w_{t}))\,p(C(w_{t})|C(w_{t-1}))
\end{equation}
where $C(\cdot)$ is a function that maps a word to its class, and $C(w_0)$ is a special start state. Brown clustering hierarchically merges clusters to maximize the quality of $C$. The quality is maximized when mutual information between all bigram classes are maximized. Although Brown clustering is commonly used, a major drawback is its limitation to learn only bigram statistics. 

\noindent\textbf{\textit{K}-means.} Because word embeddings span a semantic space, clusters of word embeddings should give a higher semantic space. We perform $K$-means on word embeddings. The resulting clusters are word classes grouped in semantic similarity under the Euclidean metric constraint. Given word embeddings $\mathbf{v}_{w_1}, \mathbf{v}_{w_2},\dots, \mathbf{v}_{w_T}$ learned from a corpus, we find the cluster membership for a word $w_t$ as
\begin{equation}
\label{eq:kmeans}
\small
k_{\mathbf{v}_{w_t}} = \arg\min_j \Arrowvert \mathbf{c}^{(j)} - \mathbf{v}_{w_t} \Arrowvert^2_2
\end{equation}
where $\mathbf{c}^{(j)}$ is the $j$th cluster centroid.
 
\subsection{Regression}
We consider linear support vector machine (SVM) regression 
\begin{equation}
\small
\min \frac{1}{2} \left \| \mathbf{w} \right \|^2 \\ \mbox{s.t.}~-\epsilon \leq y^{(i)} - (\mathbf{w}\cdot\mathbf{x}^{(i)} + b) \leq \epsilon~~\forall i
\end{equation} 
where the regressed estimate $\mathbf{w}^\top\cdot\mathbf{x}^{(i)} + b$ for $i$th input $\mathbf{x}^{(i)}$ is optimized to be bound within an error margin $\epsilon$ from the ground-truth label $y^{(i)}$. SVM trains a bias term $b$ to better compensate regression errors along the weight vector $\mathbf{w}$. We train SVM regression using feature vectors formed on word embedding and clustering to predict the readability score.

\subsection{Pipeline}
Figure~\ref{fig:ours} depicts our prediction pipeline using word clusters precomputed by K-means on word embeddings. When a document of an unknown readability level arrives, we preprocess tokenized text input and compute word vectors using trained word embeddings. We compute cluster membership on word vectors, followed by average pooling. For cluster membership, we perform the 1-of-$K$ hard assignment for each word in the document. Then we compute the histogram of cluster membership. By representing features in terms of histograms our approach can naturally address documents of varying lengths. After some post-processing (\eg, unit-normalization), we regress the readability level.

\begin{table}[t]
\centering
\caption{Baseline regression results}
\label{tab:baseline}
\footnotesize
\begin{tabular}{c||c|c}
\hline
\textbf{System} & \textbf{Spearman} & \textbf{Pearson} \\ \hline
\texttt{bag-of-words} & 0.373 & 0.433 \\ \hline
\texttt{word2vec} & 0.579 & 0.629\\ \hline
\texttt{fastText} & 0.670 & 0.639\\ \hline
\texttt{doc2vec} & 0.525 & 0.539\\ \hline
\end{tabular}
\end{table}

\begin{table}[t]
\centering
\caption{Clustering-based language model results}
\label{tab:imp1}
\footnotesize
\begin{tabular}{c||c|c}
\hline
\textbf{System} & \textbf{Spearman} & \textbf{Pearson} \\ \hline
Brown clustering & 0.546 (0.430) & 0.534 (0.443)\\ \hline
\texttt{word2vec} +$K$-means & 0.711 (0.670)& 0.705 (0.664)\\ \hline
\texttt{fastText} +$K$-means & 0.825 (0.758)& 0.822 (0.810)\\ \hline
\end{tabular}
\end{table}

\begin{table}[t]
\centering
\footnotesize
\caption{Performance comparison summary}
\label{tab:compare}
\begin{tabular}{c||c|c}
\hline
\textbf{System}            & \textbf{Spearman} & \textbf{Pearson} \\ \hline
Text length                & -                 & 0.36             \\
Flesch-Kincaid             & -                 & 0.49             \\
Flor~\etal~\cite{flor2013}         & -                 & -0.44            \\ \hline
Lexile \tablefootnote{\url{http://lexile.com}}           & 0.50              & -                \\
ATOS \tablefootnote{\url{http://www.renaissance.com}}            & 0.59              & -                \\
DRP \tablefootnote{\url{http://questarai.com}}             & 0.53              & -                \\
REAP \tablefootnote{\url{http://reap.cs.cmu.edu}}              & 0.54              & -                \\
Reading Maturity \tablefootnote{\url{http://readingmaturity.com}} & 0.69              & -                \\
SourceRater \tablefootnote{\url{http://naeptba.ets.org}}      & 0.75              & -                \\ \hline
Vajjala~\etal~\cite{vajjala2014}      & 0.69              & 0.61             \\ \hline
Our approach          & \textbf{0.83}                 & \textbf{0.82}                \\ \hline
\end{tabular}
\end{table}

\section{Experimental Evaluation}
Following Vajjala~\etal~\cite{vajjala2014}, we evaluate readability level prediction with the Common Core Standards corpus \cite{CCSC} and sentence matching with the Wiki-SimpleWiki corpus \cite{zhu2010}.

\subsection{Common Core Standards Corpus}
This corpus of 168 English excerpts are available as the Appendix B of the Common Core Standards reading initiative of the US education system. Each text excerpt is labeled with a level in five grade bands (2-3, 4-5, 6-8, 9-10, 11+) as established by educational experts. Grade levels 2.5, 4.5, 7, 9.5, and 11.5 are used as ground-truth labels. We cut the corpus into \texttt{train} and \texttt{test} sets in an uniformly random 80-20 split, resulting 136 documents for training and 32 for test.

\noindent\textbf{Evaluation metric.} For fair comparison with other work, we adopt Spearman's rank  correlation and Pearson correlation computed between the ground-truth label and regressed value.

\noindent\textbf{Preprocessing.} We convert all characters to lowercase, strip punctuations, and remove extra whitespace, URLs, currency, numbers, and stopwords using the NLTK Stopwords Corpus \cite{nltk}.

\noindent\textbf{Features.} There are two levels of features. At the word-vector level, we perform weighted average pooling of word embeddings to compose per-document feature vector. We have tried tf-idf and uniform weighting schemes. Brown clustering of \emph{words} yields the word-vector level features as well. On the contrary, $K$-means clustering of \emph{word vectors} yields higher-level features in terms of cluster structures. For Brown and $K$-means, we replace each word in a document with its numeric cluster ID and compute the histogram of cluster membership as per-document feature vector. For histogram computing, we consider binary (on/off) and traditional bin counts.  

\noindent\textbf{Word and paragraph embeddings.} We use \texttt{word2vec} for the Skip-gram word embeddings. We have first tried out the \textsc{wiki} and \textsc{ap-news} pretrained \texttt{word2vec} models. Eventually, we use TensorFlow to train \texttt{word2vec} model from the Common Core Standards corpus. We have optimized the word-vector dimension hyperparameter between 32 and 300. 

We use \texttt{fastText} for character $n$-gram word embeddings. Similar to our \texttt{word2vec} experiment, we have tried the \textsc{wiki} and \textsc{ap-news} pretrained models for \texttt{fastText} before training our own. While training, we use the negative sampling loss function with word-vector dimensions 32 to 300 and context window size of 5.

We use \texttt{doc2vec} that implements Paragraph Vector. We have not trained our own \texttt{doc2vec} model and opted for the \textsc{wiki} and \textsc{ap-news} pretrained \texttt{doc2vec} models. 

\noindent\textbf{Brown clustering.} We use an open-source implementation by Liang~\etal~\cite{liang}. We have fine-tuned the number of cluster hyperparameter by varying between 10 and 200.
 
\noindent\textbf{\textit{K}-means clustering.} After embedding all words in each document, we run $K$-means. We fine-tune $K$ within 10 to 200.

\noindent\textbf{SVM regression.} We use LIBLINEAR \cite{liblinear} for SVM regression and configure as the $\ell_2$-regularized $\ell_2$-loss linear solver with unit bias. The SVM complexity hyperparameter is optimized between $10^{-5}$ and 1. Our choice of linear SVM is made after also trying out a single-layer perceptron neural network regression with the number of neurons in 0.1x to 1x the feature vector dimension. 

\noindent\textbf{Results and discussion.} 
Our baseline results with pretrained models are shown in Table~\ref{tab:baseline}. \texttt{Bag-of-words} performs poorly, and \texttt{word2vec} performs better than \texttt{doc2vec}. We suspect that the benefit of \texttt{doc2vec} is not realized on this corpus due to its limited length. We find \texttt{fastText} superior over \texttt{word2vec} and \texttt{doc2vec}. Pretrained \textsc{wiki} outperforms \textsc{ap-news}. We only report \textsc{wiki} results.

Table~\ref{tab:imp1} presents results on clustering-based language models: Brown clustering on words and $K$-means on trained word vectors using the corpus. Presented correlation values are for binary (inside parenthesis) and traditional bin counts. While binary counters could be robust against ambiguities resulting from repeated texts in a document, this advantage is not present in the corpus we use here. Brown clustering on words has similar performance to baseline embedding schemes. The comparable performances are expected, because both Brown clustering and the baseline embedding schemes are performed on the raw words. We can improve performance further with $K$-means clustering on word vectors. Rather than training word vector models on \textsc{wiki}, training with the Common Core Standards corpus improves the correlation. \texttt{fastText} with $K$-means works the best.

Table~\ref{tab:compare} presents a summary that compares performances of our approach and the previous work. Flor~\etal~\cite{flor2013} implemented prediction scheme based on lexical tightness and compared their method against baselines such as text length and Flesch-Kincaid \cite{kincaid} in Pearson correlation. Nelson~\etal~\cite{nelson2012} wrote a summary of commercial softwares' performances in Spearman correlation. Most recently, Vajjala~\etal~\cite{vajjala2014} implemented a scheme that uses lexical, syntactic, and psycholinguistic features. Our highest correlation for Spearman is 0.83, and 0.82 for Pearson, both of which are better than the best case reported by the previous work. 

\subsection{Wiki-SimpleWiki Corpus}
 We demonstrate our features derived from clustering of word embeddings are effective in another application concerning sentence matching. The corpus for this application consists of 108,016 aligned sentence pairs of the same meaning drawn from (ordinary) Wikipedia and Simple Wikipedia.\footnote{\url{http://simple.wikipedia.org}} Simple Wikipedia uses basic vocabulary and less complex grammar to make the content of Wikipedia accessible to audiences of all reading skills.

\noindent\textbf{Task and metric.} We evaluate whether or not the feature vector for an ordinary sentence formed by the proposed feature scheme can correctly predict its counterpart sentence. We sample 1,000 sentence pairs. Among all 1,000 pairs, we compute the probability $P_N$ that ordinary sentences and their simple counterparts are $N$ nearest neighbors in the semantic space. We vary $N = 1$ to $4$. 

\noindent\textbf{Features.} We use our best feature scheme, word embedding by \texttt{fastText} and $K$-means, found in Section 3.1. To compute sentence embedding, we average-pool all word embeddings in the sentence.

\noindent\textbf{Results and discussion.} As Table~\ref{tab:wikisimple} shows, using only the nearest neighbor, we already achieve $P_N = 0.959$; as $N$ grows, we can contain different sentences of the same meaning with probability approaching 1. This implies that despite differences in grammatical structure and word usage, when underlying semantics are shared between two sentences, they are mapped closely each other in the feature space.

\begin{table}[t]
\centering
\footnotesize
\caption{Average probability $P_N$ that a Wiki sentence and its SimpleWiki counterpart are within the $N$th nearest neighbors in the semantic feature vector space}
\label{tab:wikisimple}
\begin{tabular}{c||c|c|c|c}
\hline
   & \multicolumn{4}{c}{$N$}        \\ \cline{2-5}
   & 1     & 2     & 3     & 4     \\ \hline
$P_N$ & 0.926 & 0.947 & 0.955 & 0.959 \\ \hline
\end{tabular}
\end{table}
\section{Conclusion}
Word vectors learned on neural embedding exhibit linguistic regularities and patterns explicitly. In this paper, we have introduced a regression framework on clustering-based language model using word embeddings for automatic text readability prediction. Our experiments with the Common Core Standards corpus demonstrate that features derived by clustering word embeddings are superior to classical shallow-length, bag-of-words, and other advanced features previously attempted on the corpus. We have further evaluated our approach on sentence matching using the Wiki-SimpleWiki corpus and showed that our method can effectively capture semantics even when sentences are written with different vocabulary and grammatical structures. For future work, we plan to continue our experiments with more diverse languages and larger datasets.

\footnotesize
\section{Acknowledgments}
This work is supported by the MIT Lincoln Laboratory Lincoln Scholars Program and in part by gifts from the Intel Corporation and the Naval Supply Systems Command award under the Naval Postgraduate School Agreements No.
N00244-15-0050 and No. N00244-16-1-0018. 

\bibliographystyle{ACM-Reference-Format}
\bibliography{paper} 

\end{document}